\documentclass[conference]{IEEEtran}
\IEEEoverridecommandlockouts
\usepackage{cite}
\usepackage{amsmath,amssymb,amsfonts}
\usepackage{graphicx}
\usepackage{textcomp}
\usepackage{xcolor}
\def\BibTeX{{\rm B\kern-.05em{\sc i\kern-.025em b}\kern-.08em
    T\kern-.1667em\lower.7ex\hbox{E}\kern-.125emX}}

\usepackage{diagbox}
\newcommand{\tabincell}[2]{\begin{tabular}{@{}#1@{}}#2\end{tabular}}
\usepackage{booktabs}

\usepackage{algorithm,algpseudocode}
\usepackage{amssymb}
\usepackage{mathtools}

\usepackage{xcolor}

\algnewcommand{\algorithmicforeach}{\textbf{for each}}
\algdef{SE}[FOR]{ForEach}{EndForEach}[1]
  {\algorithmicforeach\ #1\ \algorithmicdo}
  {\algorithmicend\ \algorithmicforeach}
\usepackage{algpseudocode}

\usepackage{multirow}

\begin{document}

\title{LSEBMCL: A Latent Space Energy-Based Model for Continual Learning}

\author{\IEEEauthorblockN{Xiaodi Li}
\IEEEauthorblockA{\textit{Department of Electrical and Computer Engineering} \\
\textit{The University of Texas at Dallas}\\
Richardson, USA \\
xiaodi.li@utdallas.edu}
\and
\IEEEauthorblockN{Dingcheng Li}
\IEEEauthorblockA{\textit{Vextex AI} \\
\textit{Google}\\
Kirkland, USA \\
dingchengli@google.com}
\and
\IEEEauthorblockN{Rujun Gao}
\IEEEauthorblockA{\textit{Department of Mechanical Engineering} \\
\textit{Texas A\&M University}\\
College Station, USA \\
grj1214@tamu.edu}
\and
\IEEEauthorblockN{Mahmoud Zamani}
\IEEEauthorblockA{\textit{Department of Electrical and Computer Engineering} \\
\textit{The University of Texas at Dallas}\\
Richardson, USA \\
mxz173130@utdallas.edu}
\and
\IEEEauthorblockN{Latifur Khan}
\IEEEauthorblockA{\textit{Computer Science Department} \\
\textit{The University of Texas at Dallas}\\
Richardson, USA \\
lkhan@utdallas.edu}
}

\maketitle

\begin{abstract}
Continual learning has become essential in many practical applications such as online news summaries and product classification. The primary challenge is known as catastrophic forgetting, a phenomenon where a model inadvertently discards previously learned knowledge when it is trained on new tasks. Existing solutions involve storing exemplars from previous classes, regularizing parameters during the fine-tuning process, or assigning different model parameters to each task. The proposed solution LSEBMCL (Latent Space Energy-Based Model for Continual Learning) in this work is to use energy-based models (EBMs) to prevent catastrophic forgetting by sampling data points from previous tasks when training on new ones. The EBM is a machine learning model that associates an energy value with each input data point. The proposed method uses an EBM layer as an outer-generator in the continual learning framework for NLP tasks. The study demonstrates the efficacy of EBM in NLP tasks, achieving state-of-the-art results in all experiments.
\end{abstract}


\section{Introduction}
Label prediction for continuously occurring instances is crucial in practical applications like online news summaries, product classification, and dialogue learning systems. To address these scenarios, models must acquire, fine-tune, and transfer knowledge over time, a concept referred to as continual learning \cite{parisi2019continual}. Continual Learning (CL) aims to create systems that can rapidly acquire new skills and integrate them with prior knowledge, mimicking human learning. A key challenge in CL is catastrophic forgetting, where models forget previously learned knowledge when training on new tasks.

Approaches to mitigate catastrophic forgetting can be categorized into: (1) storing exemplars from previous tasks; (2) parameter regularization during fine-tuning; and (3) task-specific parameter allocation. These methods aim to retain prior knowledge while learning new tasks. Our method prevents forgetting by sampling data from previous tasks using an Energy-based Model (EBM) during training. The EBM is first trained on each task, enabling the retention of prior knowledge and improving performance on subsequent tasks.

An EBM associates scalar energy values with input data points, assigning lower energy to more likely inputs and higher energy to less likely ones. EBMs are applicable to various tasks, including classification and regression. For instance, Pang et al. \cite{pang2020learning} used EBMs in the latent space to enhance the expressivity of generative models, demonstrating its utility in improving latent space structure for both generation and classification.

Despite advancements in CL, EBM applications in this domain remain limited. For example, \cite{li2022energy} applied EBMs for classification in complex CL scenarios like boundary-agnostic and class-incremental learning. Unlike their approach, which uses EBMs as the core model, we employ EBMs as an external sampling mechanism, simplifying updates and preserving flexibility. Additionally, while their work focuses on computer vision datasets, our study applies EBMs to natural language processing datasets.

Our proposed method, LSEBMCL, makes three key contributions: (1) integrating an EBM layer into a continual learning framework for NLP tasks for the first time; (2) addressing catastrophic forgetting in NLP tasks using EBM; and (3) achieving state-of-the-art performance across experiments, demonstrating the effectiveness of our approach.

\section{Related Works}
\subsection{Continual Learning}
Continual learning involves sequential tasks and is particularly relevant in scenarios where data arrives in a non-i.i.d. manner and new tasks emerge. However, deep neural networks face the challenge of \textit{catastrophic forgetting}, which hinders their ability to retain prior knowledge. Current continual learning methods fall into three categories: (1) Replay methods \cite{zhao2022prompt, varshney2022prompt}; (2) Regularization-based methods \cite{li2022overcoming, li2022continual}; and (3) Parameter isolation methods \cite{zhu2022continual, qin2022lfpt}. Replay methods either store raw samples or generate pseudo-samples using generative models. For instance, iCaRL \cite{rebuffi2017icarl} stores class exemplars, and GEM \cite{lopez2017gradient} uses gradients from previous tasks to constrain updates. LAMOL \cite{sun2019lamol} reduces forgetting by generating artificial examples of previous tasks. Regularization-based methods avoid storing raw inputs, introducing regularization terms to consolidate knowledge. LwF \cite{li2017learning} uses model outputs as soft labels, while MAS \cite{aljundi2018memory} estimates parameter importance for adaptation. IDBR \cite{huang2021continual} applies disentanglement-based regularization for text classification, and LPC \cite{li2022lpc} combines parameter calibration with logit preservation. Parameter isolation methods allocate distinct parameters per task to prevent forgetting, as seen in PackNet \cite{mallya2018packnet} and HAT \cite{serra2018overcoming}. Our method employs replay-based EBMs to generate artificial samples for previous tasks.

\subsection{Energy-based Models}
Energy-based models (EBMs) represent probability density functions via an energy function $E(x)$, mapping realistic points to low energy values and unrealistic points to high values \cite{lecun2006tutorial}. EBMs offer simplicity, stability, and parameter efficiency. Recent advancements enable EBMs to model high-dimensional data \cite{xie2016theory, nijkamp2019learning}, and latent space EBMs \cite{pang2020learning} improve model expressivity for tasks such as text generation and trajectory modeling. EBMs have been shown to prevent catastrophic forgetting in continual learning \cite{li2022energy}. Unlike \cite{li2022energy}, which uses EBMs as the primary model for classification, our method employs EBMs as an outer-generator for tasks such as classification and text generation. Other applications of EBMs include joint training with pretrained text encoders to enhance calibration \cite{he2021joint} and leveraging low-dimensional structures for anomaly detection \cite{yoon2023energy}.

\section{Methodology}
\begin{figure*}[ht]
    \centering
    \includegraphics[width=6.3in]{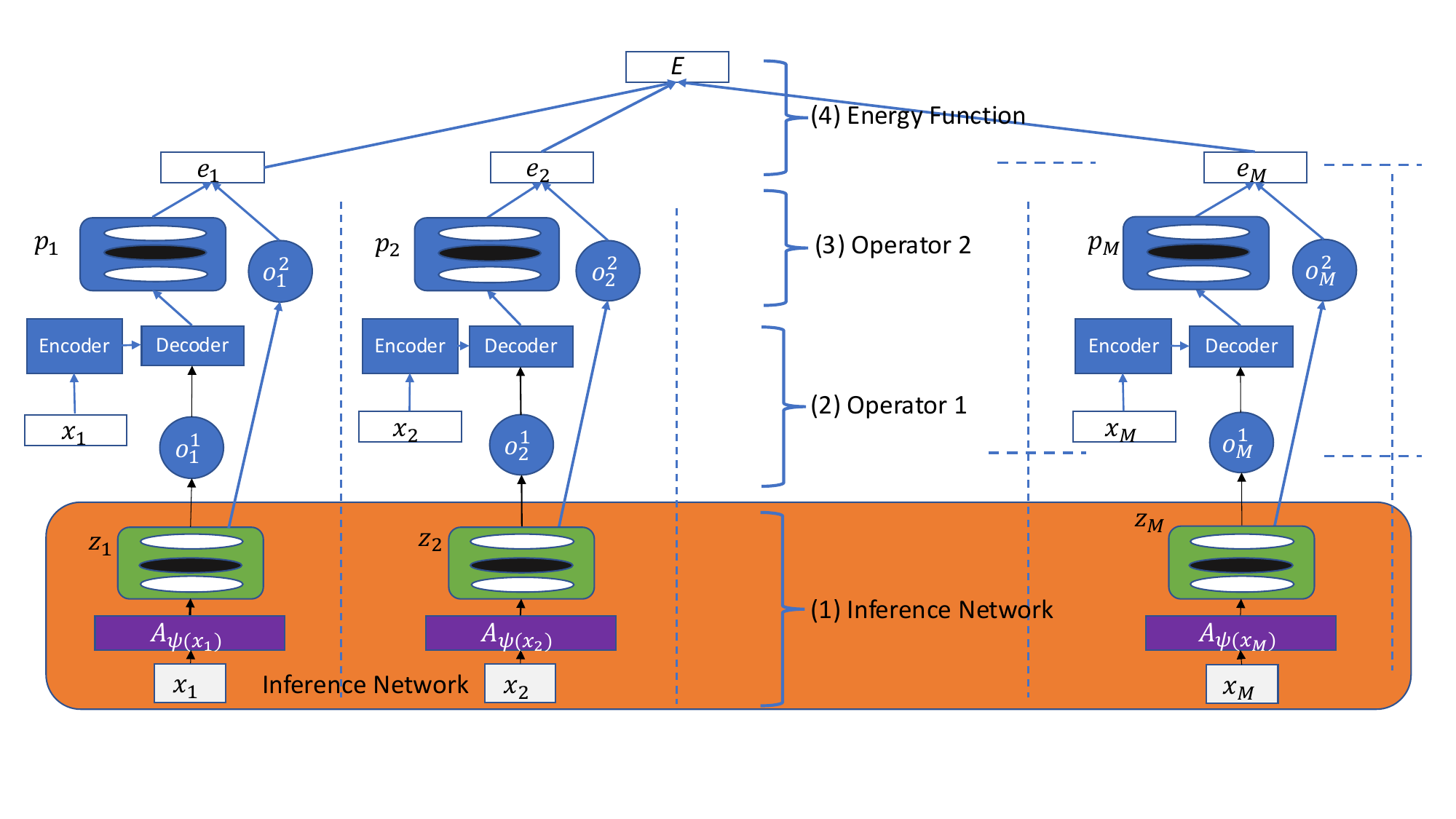}
    \caption{The Overview of LSEBMCL Framework. (1) Inference Network: The process begins with the inference network at the bottom, where inputs (x) are processed to generate encoded representations (z). (2) Operator 1 and (3) Operator 2: These operators facilitate the transition of logits from the inference network to the decoder inputs and compute the energy on the outputs, respectively. (4) Energy Function: At the culmination of the process, the energy function evaluates the outputs, contributing to the model's generation.}
    \label{fig:Overview}
\end{figure*}
We initiate the process with a pre-trained base model, specifically the latest Large Language Model (LLM) Mistral 7B. \cite{jiang2023mistral}. We have developed the LSEBMCL model, which consists of four main components. The initial component is the Inference Network, followed by Operator 1, Operator 2, and finally, the Energy Function. In the subsequent sections, we will provide a detailed introduction to each of these components individually.

\subsection{Inference Network}
Following decaNLP \cite{mccann2018natural}, we make the datasets pre-processed as a QA (Question Answering) problem. First, we convert continual learning tasks to a unified QA format as:

\begin{equation}
    <x, y> \in D
\end{equation}

\noindent where x is the Question, y is the Answer, and D is a training set of QA. We target handling diverse tasks, covering Question Answering (QA), Natural Language Inference (NLI), Sentiment Analysis (SA), Semantic Role Labeling (SRL), etc.

We introduce an inference network, denoted as $A_{\Psi}(x)$, which is also referred to as the "energy-based inference network" (as depicted at the bottom of Figure \ref{fig:Overview}). This network is parameterized by $\Psi$ and trained with the objective of achieving the following goal:

\begin{equation}
    A_{\Psi}(x) \approx argmin_{y\in Y_{R}(x)}E_{\Theta}(x, y)
\end{equation}

\noindent where $Y_{R}(x)$ are the true labels. Specifically, we train the inference network parameters $\Psi$ as follows (assuming $\Theta$ is the parameters of EBM):

\begin{equation}
    \widehat{\Psi} = argmin_{\Psi}\sum_{<x, y>\in D}E_{\Theta}(x, A_{\Psi}(x))
\end{equation}

\subsection{Operator 1}
We employ two operators $o^{1}$ and $o^{2}$ that are used to map $z_{t}$ logits into distributions for use in the energy. As shown at the middle of Figure \ref{fig:Overview}, we seek an operator $o^{1}$ to modulate the way that logits $z_{t}$ output by the inference network is fed to the decoder input slots in the energy function. $o^{1}$ is the operation for feeding inference network outputs into the decoder input slots in the energy.

\subsection{Operator 2}
An operator $o^{2}$ to determine how the distribution $p_{\Theta}(\cdot| ...)$ is used to compute the log probability of $y$. $o^{2}$ is the operation for computing the energy on the output. Explicitly, then, we write each local energy term as

\begin{equation}
    e_{m}(x,y) = -o^{2}(z_{m})\log p_{\Theta}(\cdot|o^{1}(z_{m}), x_{m})
\end{equation}

Our objective is to minimize the aforementioned energy function concerning the variable $z_t$ in our inference networks. The softmax operation is selected for $o^1$ and $o^2$.

\subsection{Energy Function}
Figure \ref{fig:Overview} shows the proposed latent space EBM continual learning model. We design an EBM layer to play the role of outer-generator. After training each task, the EBM model generates samples based on data from previous tasks before training on the new task. During training on the new task, the model not only trains a model on the new data but also trains on the extra data generated from previous tasks by the EBM model as shown at the top of Figure \ref{fig:Overview}. The equation is shown below:

\begin{equation}
    E_{\Theta}(x, y) = \sum_{m=1}^{M}e_{m}(x, y)
\end{equation}

\noindent where $\Theta$ is the parameters of the EBM. $m$ means the index of the task and $M$ means the total number of tasks. $e_{m}(x, y)$ is calculated using the following equation:

\begin{equation}
    e_{m}(x, y)=-\log p(y_{m}|x_{m})
\end{equation}

In Figure \ref{fig:Overview}, for each task, we have:

\begin{equation}
    p_{\theta}(x, z) = p_{\alpha}(z)p_{\beta}(x|z)
\end{equation}

\noindent where $p_{\alpha}(z)$ is the prior model with parameters $\alpha$, $z$ is a latent dense continuous vector, and $p_{\beta}(x|z)$ is given by a generative model parameterized with $\beta$. 

The EBM prior of $z$, $p_{\alpha}(z)$, is in the form of the energy-based correction of an isotropic Gaussian reference distribution $p_{0}(z)$:

\begin{equation}
\begin{aligned}
    p_{\alpha}(z) = \frac{1}{Z(\alpha)}\exp[F_{\alpha}(z)]p_{0}(z)
    &\propto\exp[F_{\alpha}(z) - \frac{1}{2\sigma^{2}}\Vert z\Vert^{2}] \\
    &=\exp E(z)
\end{aligned}
\end{equation}

\noindent where $E(z) = F_{\alpha}(z) - \frac{1}{2\sigma^{2}}\Vert z\Vert^{2}$ is the energy function that maps $z$ to a scalar,
where $F_{\alpha}(z)$ is parameterized by a small multi-layer perceptron (MLP), $\sigma^{2}$ is the regularization related hyper-parameter. $Z(\alpha) = \int\exp[F_{\alpha}(z)]p_{0}(z)dz$ is the normalizing constant/partition function.

For language modeling, assuming $x\in R^{D}$,
\begin{equation}
    x = g_{\beta}(z) + \epsilon
\end{equation}

\noindent where $\epsilon\sim N(0, \sigma^{2}I_{D})$, so that $p_{\beta}(x|z)\sim N(g_{\beta}(z), \sigma^{2}I_{D})$. $g_{\beta}(z)$ can be a transformer, bert, gpt, bart for language learning.

For text modeling, $p_{\beta}(x|z)$ is an autoregressive model as shown in equation \ref{equ_autoregressive}: 

\begin{equation}\label{equ_autoregressive}
    p_{\beta}(x|z)=\prod_{t}^{T}p_{\beta}(x^{(t)}|x^{(1)}, ..., x^{(t-1)}, z)
\end{equation}

Let $p_{data}(x)$ be the data distribution, learning of $\beta$ of $p_{\beta}(x)$ can be based on $\min_{\beta}KL(p_{data}(x)|p_{\beta(x)})$.

Observe $\{x_{i}, i=1, ..., N\}\sim p_{data}(x)$,

\begin{equation}
\begin{aligned}
    \min_{\beta}KL(p_{data}(x)|p_{\beta}(x))&\approx L(\beta)=\frac{1}{N}\sum_{i}^{N}\log p_{\beta}(x_{i})
\end{aligned}
\end{equation}

\noindent which is the Maximum Likelihood Estimation (MLE). We calculate the gradient of $\log p_{\beta}(x)$:

\begin{equation}
\label{equ:gradient}
\begin{aligned}
    \nabla_{\beta}\log p_{\beta}(x) &= \frac{1}{p_{\beta}(x)}\nabla_{\beta}p_{\beta}(x) \\ &= \frac{1}{p_{\beta}(x)}\int \nabla_{\beta}p_{\beta}(x, z)dz \\ &= E_{p_{\beta}(z|x)}[\nabla_{\beta}\log p_{\beta}(x, z)]
\end{aligned}
\end{equation}

The marginal distribution of $x$ is $p_{\beta}(x)=\int p_{\beta}(x, z)dz$. The inference of z based on posterior is $p_{\beta}(z|x)=p_{\beta}(x, z)/p_{\beta}(x)$. 

With gradient descent, in each iteration, we have:

\begin{equation}
\begin{aligned}
    \beta_{t+1} = \beta_{t} + \delta_{t}\frac{1}{N}E_{p_{\beta_{t}}(z_{i}|x_{i})}[\nabla\log p_{\beta}(x_{i}, z_{i})|\beta = \beta_{t}]
\end{aligned}
\end{equation}

\noindent where $E_{p_{\beta_{t}}(z_{i}|x_{i})}[...]$ is the gradient calculated by equation \ref{equ:gradient}, approximated by short-run MCMC inference dynamics with Langevin dynamics.

We approximate the intractable expectation $E_{p}$ with MCMC, Langevin dynamics, a gradience-based MCMC. We draw samples from the EBM prior:

\begin{equation}
\begin{aligned}
    z_{k + 1} = z_{k} - s\nabla_{z}\log p_{\alpha}(z_{k}) + \sqrt{2s}\epsilon_{k}, \\
    k = 1, ..., K, z_{0}\sim p_{0}(z), \epsilon_{k}\sim N(0, I_{d})
\end{aligned}
\end{equation}

The short-run Langevin dynamics is always initialized from the fixed initial distribution $p_{0}$, and only runs a fixed number of $K$ steps, e.g., $K = 20$. 

Similarly, we can also draw samples from the posterior distribution $p_{\beta}(z|x)$:

\begin{equation}
\begin{aligned}
    z_{k + 1} = z_{k} - s\nabla_{z}\log p_{\beta}(z_{k}|x) + \sqrt{2s}\epsilon_{k}, \\
    k = 1, ..., K, z_{0}\sim p_{0}(z), \epsilon_{k}\sim N(0, I_{d})
\end{aligned}
\end{equation}

\noindent where $s$ is the small Langevin step size, $t$ indexes the time step of the Langevin dynamics, $\nabla_{z}\log p_{\alpha}(z_{k})$ or $\nabla_{z}\log p_{\beta}(z_{k}|x)$ can be efficiently computed by back propagation.

\section{Experiments}
In this section, we assess the performance of our model on various tasks. In our experiment, we focus on task-incremental learning. We conduct a comparison of our method with eleven different techniques. All the methods use Mistral 7B as the backbone pretrained large language model. In the continual learning scenario, we train the model sequentially on a series of distinct tasks, following a predetermined order. After each training phase, we assess the model's performance on all previously encountered tasks.
\begin{table*}
\centering
\caption{Summary of tasks, datasets, dataset sizes, and their corresponding metrics. As this work uses no development set, only the training and test datasets are shown. nF1 is the normalized version of the F1 score; EM represents an exact match between texts: for text classification, this amounts to accuracy; for WOZ, it is equivalent to dfEM (turn-based dialogue state exact match); for WikiSQL, it is equivalent to lfEM (exact match of logical forms).}
    \begin{tabular}{|l|l|r|r|r|}
         \hline
         \textbf{Task} & \textbf{Dataset} & \textbf{\# Train} & \textbf{\# Test} & \textbf{Metric} \\
         \hline
         Question answering & SQuAD 2.0 & 130319 & 11873 & nF1 \\
         \hline
         Semantic parsing & WikiSQL & 56355 & 15878 & lfEM \\
         \hline
         Sentiment analysis & SST & 6920 & 1821 & EM \\
         \hline
         Semantic role labeling & QA-SRL & 6414 & 2201 & nF1 \\
         \hline
         Goal-oriented dialogue & WOZ & 2536 & 1646 & dsEM \\
         \hline
         \multirow{5}{*}{Text classification} & AGNews & \multirow{5}{*}{115000} & \multirow{5}{*}{7600} & \multirow{5}{*}{EM} \\
         \multirow{5}{*}{} & Amazon & \multirow{5}{*}{} & \multirow{5}{*}{} & \multirow{5}{*}{} \\
         \multirow{5}{*}{} & DBPedia & \multirow{5}{*}{} & \multirow{5}{*}{} & \multirow{5}{*}{} \\
         \multirow{5}{*}{} & Yahoo & \multirow{5}{*}{} & \multirow{5}{*}{} & \multirow{5}{*}{} \\
         \multirow{5}{*}{} & Yelp & \multirow{5}{*}{} & \multirow{5}{*}{} & \multirow{5}{*}{} \\
         \hline
    \end{tabular}
    \label{tab:data}
\end{table*}


\begin{table*}
\caption{Summary of averaged metric scores for different methods under permuted task orders using models at the last epoch of the last task. The Average and Std columns respectively are the average and standard deviation of the averaged scores for each row of the methods. Multitasked learning as an upper bound is shown at the bottom.}
  \resizebox{\textwidth}{25mm}{
  \begin{tabular}{|l|c|c|c|c|c|c|c|c|}
    \hline
    \textbf{Model} & \tabincell{c}{\textbf{SST SRL WOZ}} &
    \tabincell{c}{\textbf{SST WOZ SRL}} &
    \tabincell{c}{\textbf{SRL SST WOZ}} & \tabincell{c}{\textbf{SRL WOZ SST}} & 
    \tabincell{c}{\textbf{WOZ SST SRL}} &
    \tabincell{c}{\textbf{WOZ SRL SST}} & \tabincell{c}{\textbf{Average}} & \tabincell{c}{\textbf{Std}} \\ 
    \hline
    Fine-tuned & 52.0 & 25.5 & 64.6 & 33.2 & 34.6 & 34.6 & 40.8 & 14.6 \\
    \hline
    EWC & 51.0 & 50.0 & 66.4 & 37.5 & 44.8 & 40.7 & 48.4 & 10.2 \\
    \hline
    MAS & 37.3 & 46.2 & 57.3 & 32.3 & 50.3 & 32.2 & 42.6 & 10.3 \\
    \hline
    GEM & 52.1 & 31.5 & 64.0 & 32.9 & 45.3 & 36.7 & 43.8 & 12.6 \\
    \hline
    LAMOL $_{GEN}^{0}$ & 47.1 & 38.4 & 57.8 & 39.4 & 45.5 & 46.6 & 45.8 & 7.0 \\
    \hline
    LAMOL $_{GEN}^{0.05}$ & 81.5 & 79.5 & 74.8 & 73.7 & 70.2 & 75.8 & 75.9 & 4.1 \\
    \hline
    LAMOL $_{GEN}^{0.2}$ & 80.8 & 81.1 & 81.2 & 80.3 & 78.9 & 81.9 & 80.7 & 1.0 \\
    \hline
    RVAE-LAMOL $_{GEN}^{0.05}$ & 80.7 & 79.6 & 79.9 & 80.2 & 79.6 & 78.0 & 79.7 & 0.9 \\
    \hline
    RVAE-LAMOL $_{GEN}^{0.2}$ & 81.5 & 82.3 & 81.5 & 82.0 & 81.1 & 82.8 & 81.9 & 0.6 \\
    \hline
    LSEBMCL $_{GEN}^{0}$ & 66.4 & 57.7 & 77.7 & 66.8 & 64.6 & 65.6 & 66.5 & 6.4 \\
    \hline
    LSEBMCL $_{GEN}^{0.05}$ & 82.8 & 81.8 & \textbf{84.4} & 81.9 & 82.7 & 80.5 & 82.4 & 1.3 \\
    \hline
    LSEBMCL $_{GEN}^{0.2}$ & \textbf{83.1} & \textbf{82.5} & 82.7 & \textbf{82.4} & \textbf{83.7} & \textbf{83.2} & \textbf{82.9} & \textbf{0.5} \\
    \hline
    Multitasked & \multicolumn{8}{c|}{87.0} \\
    \hline
    \end{tabular}}
    \label{tab:results}
\end{table*}

\begin{table*}
\caption{Summary of the averaged score on five tasks. The sequence order is SQuAD 2.0,  WikiSQL, SST, QA-SRL, and WOZ. The scores are reported as the averaged score over all tasks of the models after training on every task. The rightmost column Multitasked is the upper bound for comparison. The best performance is in boldface.}
    \resizebox{\textwidth}{4.5mm}{
    \begin{tabular}{|c|c|c|c|c|c|c|c|c|}
        \hline
        \textbf{Fine-tuned} & \textbf{MAS} & \textbf{LAMOL $_{GEN}^{0.05}$} & \textbf{LAMOL $_{GEN}^{0.2}$} & \textbf{HMI-LAMOL$_{GEN}^{0.05}$} & \textbf{HMI-LAMOL$_{GEN}^{0.2}$} & \textbf{LSEBMCL $_{GEN}^{0.05}$} & \textbf{LSEBMCL $_{GEN}^{0.2}$} & \textbf{Multitasked}\\
        \hline
        52.6 & 51.2 & 70.3 & 74.1 & 76.0 & 76.9 & \textbf{76.5} & \textbf{77.3} & 78.2 \\ 
        \hline
    \end{tabular}}
    \label{tab:fivetasks}
\end{table*}

\begin{table*}
\caption{Summary of results on text classification tasks using averaged EM score (equivalent to averaged accuracy in \cite{de2019episodic}) of models at last epoch of last task. The four orders mirror those in \cite{de2019episodic}. For MBPA++, MBPA++ (our impl.), LAMOL $_{TASK}^{0.2}$, PMR, IDBR, ProgPrompt, HMI-LAMOL, and LSEBMCL$^{0.05}_{GEN}$, the results are averaged over two runs.}
    \centering
    \resizebox{\textwidth}{13mm}{
    \begin{tabular}{|c|c|c|c|c|c|c|c|c|}
        \hline
        \textbf{Order} & \textbf{MBPA++} & \textbf{MBPA++ (our impl.)} & \textbf{LAMOL $_{TASK}^{0.2}$} & \textbf{PMR} & \textbf{IDBR} & \textbf{ProgPrompt} & \textbf{HMI-LAMOL} & \textbf{LSEBMCL $_{GEN}^{0.05}$} \\
        \hline
        i & 70.8 & 75.3 & 78.6 & 73.5 & 77.0 & 78.9 & 77.8 & \textbf{80.2} \\
        \hline
        ii & 70.9 & 76.0 & 78.3 & 74.3 & 77.2 & 78.4 & 78.7 & \textbf{80.0} \\
        \hline
        iii & 70.2 & 73.9 & 78.0 & 72.0 & 78.1 & 79.0 & 79.7 & \textbf{80.2}\\
        \hline
        iv & 70.7 & 76.7 & 76.9 & 70.2 & 77.8 & 78.0 & 78.4 & \textbf{80.3} \\
        \hline
        Average & 70.7 & 75.5 & 77.9 & 72.5 & 77.5 & 79.0 & 78.6 & \textbf{80.2} \\
        \hline
    \end{tabular}}
    \label{tab:classification}
\end{table*}

\subsection{Experimental Setup}
\subsubsection{Tasks, Datasets, and Metrics}
We collect datasets for five different tasks related to natural language processing mentioned in decaNLP \cite{mccann2018natural}, including question answering, semantic parsing, sentiment analysis, semantic role labeling, and goal-oriented dialogue. To compare our method with \cite{de2019episodic}, we also conduct experiments on four text classification tasks: news classification, sentiment analysis, Wikipedia article classification, and question-and-answer categorization with five datasets, following the same procedure for producing equal-sized datasets. Due to limited computational resources, we did not train on all datasets. We use a corresponding evaluation metric for each task. Table \ref{tab:data} summarizes the tasks, datasets, and metrics. The scores for the metrics range between 0 and 100\%.

\subsubsection{Methods for Comparison}
The paper discusses various approaches to tackling the problem of catastrophic forgetting in continual learning, where a model trained on a sequence of tasks tends to forget the previous tasks when trained on subsequent tasks. The approaches considered in the paper include fine-tuning, multi-task learning, replay methods and architecture-based methods LAMOL, RVAE-LAMOL \cite{wang2022rvae}, HMI-LAMOL \cite{maekawa2023generative}, PMR \cite{ho2023prototype}, regularization-based methods such as Online EWC \cite{schwarz2018progress} and MAS \cite{aljundi2018memory}, Gradient Episodic Memory (GEM) \cite{lopez2017gradient}, Improved Memory-Based Parameter Adaptation (MBPA++) \cite{de2019episodic}, IDBR \cite{huang2021continual}, and other methods like ProgPrompt \cite{razdaibiedina2023progressive}:

(1) \textbf{LSEBMCL}: Uses top-$k$ sampling with $k=1$. LSEBMCL$^{\gamma}_{GEN}$ denotes a sampling ratio $\gamma$, applying the same GEN token across tasks. (2) \textbf{LAMOL}: Employs $k=20$ for top-$k$ sampling and $\lambda=0.25$ for LM loss weighting. (3) \textbf{RVAE-LAMOL}: Enhances LAMOL with a residual variational autoencoder. (4) \textbf{HMI-LAMOL}: Adds hippocampal memory indexing to improve generative replay via compressed features. (5) \textbf{PMR}: Stores minimal samples for efficient continual learning. (6) \textbf{Fine-tuning}: Sequentially trains tasks without task interaction. (7) \textbf{Multitask learning}: Trains all tasks simultaneously, serving as a continual learning upper bound. (8) \textbf{Regularization methods}: Includes Online EWC and MAS for mitigating forgetting. (9) \textbf{GEM}: Randomly samples 5\% of prior task data for gradient calculation. (10) \textbf{MBPA++}: Combines sparse experience replay with local adaptation. (11) \textbf{IDBR}: Uses disentanglement-based regularization for text classification. (12) \textbf{ProgPrompt}: Prevents catastrophic forgetting without data replay or extensive task-specific parameters.

\subsection{Experimental Results}

\subsubsection{SST, QA-SRL, and WOZ Tasks}
To gain a preliminary understanding of the effectiveness of the different methods and the impact of the task order, we conducted an experiment on three small datasets: SST, QA-SRL, and WOZ. We trained all methods except for the multitasked method on six different orders of tasks. We evaluated the model's final score after training on each order, and the results are presented in Table \ref{tab:results}. Based on the results, we made several observations. We observed several things as follows: (1) Fine-tuned, EWC, MAS, GEM, LAMOL, and RVAE-LAMOL had worse performance than LSEBMCL even with $\gamma = 0$ and significantly worse than LSEBMCL with $\gamma > 0$. (2) LSEBMCL$^{0.2}_{GEN}$ achieves the best performance, even approximating the multitasked upper bound with 2.6\%, implying little forgetting during continual learning. (3) Task order does influence performance with LSEBMCL. (4) When using LSEBMCL, the performance of old tasks remained almost the same throughout the training process. Increasing the sampling ratio $\gamma$ improved the performance, particularly when increased from 0 to 0.05. (5) A better continual method had a smaller standard deviation, indicating it was less affected by the task order. LSEBMCL even achieves the lowest standard deviation among all the baselines, indicating its robustness to task order variations.

\subsubsection{Five DecaNLP Tasks}
In this sequential training experiment, five tasks were tackled in order of decreasing size, commencing with the largest task (SQuAD 2.0) and concluding with the smallest (WOZ). This task sequence was determined by the constraints of limited computing resources. Notably, LSEBMCL exhibited superior performance across all tasks, outperforming other methods by a significant margin and even approximates the multitasked upper bound with 0.9\%, as detailed in Table \ref{tab:fivetasks}. Moreover, the effectiveness of LSEBMCL demonstrated further enhancement with an increase in the sampling ratio $\gamma$. The experiment's results underscore LSEBMCL's remarkable efficacy and suitability for diverse tasks, confirming its robust performance.

\subsubsection{Text Classification Tasks}
We compare our proposed method, LSEBMCL, against the state-of-the-art MBPA++, LAMOL, PMR, IDBR, ProgPrompt, and HMI-LAMOL. The results are shown in Table \ref{tab:classification}. LSEBMCL$^{0.05}_{GEN}$ outperformed LAMOL$^{0.2}_{TASK}$, our implementation of MBPA++, PMR, IDBR, ProgPrompt, and HMI-LAMOL even with sampling ratio 0.05 and GEN token. This indicates that the improvements made in LSEBMCL were significant and that it is a strong method for mitigating catastrophic forgetting with less sampling data.


\section{Conclusion}
In this study, we introduce an innovative approach known as LSEBMCL that integrates an EBM layer into the continual learning framework for NLP tasks. In addition to its promising applications in NLP tasks, it holds potential implications for computer vision tasks. Leveraging the expressive power of the EBM prior in text modeling, we construct a latent space conducive to interpretable generation and text classification. To achieve this, we devise a novel prior distribution that integrates continuous latent variables for generation and discrete latent variables for inducing structural elements. Furthermore, we utilize the EBM to generate samples from previous tasks when training the model on new tasks. The experiments demonstrate the superior performance of our proposed approach. 

\section*{Acknowledgement}
The research reported herein was supported in part by NSF awards DMS-1737978, DGE-2039542, OAC-1828467, OAC-1931541, and DGE-1906630, ONR awards N00014-17-1-2995 and N00014-20-1-2738, Army Research Office Contract No. W911NF2110032 and IBM faculty award (Research). 

\bibliographystyle{IEEEtran}
\bibliography{IEEEabrv}

\begin{thebibliography}{10}
\providecommand{\url}[1]{#1}
\csname url@samestyle\endcsname
\providecommand{\newblock}{\relax}
\providecommand{\bibinfo}[2]{#2}
\providecommand{\BIBentrySTDinterwordspacing}{\spaceskip=0pt\relax}
\providecommand{\BIBentryALTinterwordstretchfactor}{4}
\providecommand{\BIBentryALTinterwordspacing}{\spaceskip=\fontdimen2\font plus
\BIBentryALTinterwordstretchfactor\fontdimen3\font minus \fontdimen4\font\relax}
\providecommand{\BIBforeignlanguage}[2]{{%
\expandafter\ifx\csname l@#1\endcsname\relax
\typeout{** WARNING: IEEEtran.bst: No hyphenation pattern has been}%
\typeout{** loaded for the language `#1'. Using the pattern for}%
\typeout{** the default language instead.}%
\else
\language=\csname l@#1\endcsname
\fi
#2}}
\providecommand{\BIBdecl}{\relax}
\BIBdecl

\bibitem{parisi2019continual}
G.~I. Parisi, R.~Kemker, J.~L. Part, C.~Kanan, and S.~Wermter, ``Continual lifelong learning with neural networks: A review,'' \emph{Neural Networks}, vol. 113, pp. 54--71, 2019.

\bibitem{pang2020learning}
B.~Pang, T.~Han, E.~Nijkamp, S.-C. Zhu, and Y.~N. Wu, ``Learning latent space energy-based prior model,'' \emph{Advances in Neural Information Processing Systems}, vol.~33, pp. 21\,994--22\,008, 2020.

\bibitem{li2022energy}
S.~Li, Y.~Du, G.~van~de Ven, and I.~Mordatch, ``Energy-based models for continual learning,'' in \emph{Conference on Lifelong Learning Agents}.\hskip 1em plus 0.5em minus 0.4em\relax PMLR, 2022, pp. 1--22.

\bibitem{zhao2022prompt}
Y.~Zhao, Y.~Zheng, Z.~Tian, C.~Gao, B.~Yu, H.~Yu, Y.~Li, J.~Sun, and N.~L. Zhang, ``Prompt conditioned vae: Enhancing generative replay for lifelong learning in task-oriented dialogue,'' \emph{arXiv preprint arXiv:2210.07783}, 2022.

\bibitem{varshney2022prompt}
V.~Varshney, M.~Patidar, R.~Kumar, L.~Vig, and G.~Shroff, ``Prompt augmented generative replay via supervised contrastive learning for lifelong intent detection,'' in \emph{Findings of the Association for Computational Linguistics: NAACL 2022}, 2022, pp. 1113--1127.

\bibitem{li2022overcoming}
D.~Li, Z.~Chen, E.~Cho, J.~Hao, X.~Liu, F.~Xing, C.~Guo, and Y.~Liu, ``Overcoming catastrophic forgetting during domain adaptation of seq2seq language generation,'' in \emph{Proceedings of the 2022 Conference of the North American Chapter of the Association for Computational Linguistics: Human Language Technologies}, 2022, pp. 5441--5454.

\bibitem{li2022continual}
G.~Li, Y.~Zhai, Q.~Chen, X.~Gao, J.~Zhang, and Y.~Zhang, ``Continual few-shot intent detection,'' in \emph{Proceedings of the 29th International Conference on Computational Linguistics}, 2022, pp. 333--343.

\bibitem{zhu2022continual}
Q.~Zhu, B.~Li, F.~Mi, X.~Zhu, and M.~Huang, ``Continual prompt tuning for dialog state tracking,'' \emph{arXiv preprint arXiv:2203.06654}, 2022.

\bibitem{qin2022lfpt}
\BIBentryALTinterwordspacing
C.~Qin and S.~Joty, ``{LFPT}5: A unified framework for lifelong few-shot language learning based on prompt tuning of t5,'' in \emph{International Conference on Learning Representations}, 2022. [Online]. Available: \url{https://openreview.net/forum?id=HCRVf71PMF}
\BIBentrySTDinterwordspacing

\bibitem{rebuffi2017icarl}
S.-A. Rebuffi, A.~Kolesnikov, G.~Sperl, and C.~H. Lampert, ``icarl: Incremental classifier and representation learning,'' in \emph{Proceedings of the IEEE conference on Computer Vision and Pattern Recognition}, 2017, pp. 2001--2010.

\bibitem{lopez2017gradient}
D.~Lopez-Paz and M.~Ranzato, ``Gradient episodic memory for continual learning,'' \emph{Advances in neural information processing systems}, vol.~30, pp. 6467--6476, 2017.

\bibitem{sun2019lamol}
F.-K. Sun, C.-H. Ho, and H.-Y. Lee, ``Lamol: Language modeling for lifelong language learning,'' \emph{arXiv preprint arXiv:1909.03329}, 2019.

\bibitem{li2017learning}
Z.~Li and D.~Hoiem, ``Learning without forgetting,'' \emph{IEEE transactions on pattern analysis and machine intelligence}, vol.~40, no.~12, pp. 2935--2947, 2017.

\bibitem{aljundi2018memory}
R.~Aljundi, F.~Babiloni, M.~Elhoseiny, M.~Rohrbach, and T.~Tuytelaars, ``Memory aware synapses: Learning what (not) to forget,'' in \emph{Proceedings of the European Conference on Computer Vision (ECCV)}, 2018, pp. 139--154.

\bibitem{huang2021continual}
Y.~Huang, Y.~Zhang, J.~Chen, X.~Wang, and D.~Yang, ``Continual learning for text classification with information disentanglement based regularization,'' \emph{arXiv preprint arXiv:2104.05489}, 2021.

\bibitem{li2022lpc}
X.~Li, Z.~Wang, D.~Li, L.~Khan, and B.~Thuraisingham, ``Lpc: A logits and parameter calibration framework for continual learning,'' in \emph{Findings of the Association for Computational Linguistics: EMNLP 2022}, 2022, pp. 7142--7155.

\bibitem{mallya2018packnet}
A.~Mallya and S.~Lazebnik, ``Packnet: Adding multiple tasks to a single network by iterative pruning,'' in \emph{Proceedings of the IEEE conference on Computer Vision and Pattern Recognition}, 2018, pp. 7765--7773.

\bibitem{serra2018overcoming}
J.~Serra, D.~Suris, M.~Miron, and A.~Karatzoglou, ``Overcoming catastrophic forgetting with hard attention to the task,'' in \emph{International Conference on Machine Learning}.\hskip 1em plus 0.5em minus 0.4em\relax PMLR, 2018, pp. 4548--4557.

\bibitem{lecun2006tutorial}
Y.~LeCun, S.~Chopra, R.~Hadsell, M.~Ranzato, and F.~Huang, ``A tutorial on energy-based learning,'' \emph{Predicting structured data}, vol.~1, no.~0, 2006.

\bibitem{xie2016theory}
J.~Xie, Y.~Lu, S.-C. Zhu, and Y.~Wu, ``A theory of generative convnet,'' in \emph{International Conference on Machine Learning}.\hskip 1em plus 0.5em minus 0.4em\relax PMLR, 2016, pp. 2635--2644.

\bibitem{nijkamp2019learning}
E.~Nijkamp, M.~Hill, S.-C. Zhu, and Y.~N. Wu, ``Learning non-convergent non-persistent short-run mcmc toward energy-based model,'' \emph{Advances in Neural Information Processing Systems}, vol.~32, 2019.

\bibitem{he2021joint}
T.~He, B.~McCann, C.~Xiong, and E.~Hosseini-Asl, ``Joint energy-based model training for better calibrated natural language understanding models,'' \emph{arXiv preprint arXiv:2101.06829}, 2021.

\bibitem{yoon2023energy}
S.~Yoon, Y.-U. Jin, Y.-K. Noh, and F.~C. Park, ``Energy-based models for anomaly detection: A manifold diffusion recovery approach,'' \emph{arXiv preprint arXiv:2310.18677}, 2023.

\bibitem{jiang2023mistral}
A.~Q. Jiang, A.~Sablayrolles, A.~Mensch, C.~Bamford, D.~S. Chaplot, D.~d.~l. Casas, F.~Bressand, G.~Lengyel, G.~Lample, L.~Saulnier \emph{et~al.}, ``Mistral 7b,'' \emph{arXiv preprint arXiv:2310.06825}, 2023.

\bibitem{mccann2018natural}
B.~McCann, N.~S. Keskar, C.~Xiong, and R.~Socher, ``The natural language decathlon: Multitask learning as question answering,'' \emph{arXiv preprint arXiv:1806.08730}, 2018.

\bibitem{de2019episodic}
C.~de~Masson~D'Autume, S.~Ruder, L.~Kong, and D.~Yogatama, ``Episodic memory in lifelong language learning,'' \emph{Advances in Neural Information Processing Systems}, vol.~32, 2019.

\bibitem{wang2022rvae}
H.~Wang, R.~Fu, X.~Zhang, and J.~Zhou, ``Rvae-lamol: Residual variational autoencoder to enhance lifelong language learning,'' in \emph{2022 International Joint Conference on Neural Networks (IJCNN)}.\hskip 1em plus 0.5em minus 0.4em\relax IEEE, 2022, pp. 1--9.

\bibitem{maekawa2023generative}
A.~Maekawa, H.~Kamigaito, K.~Funakoshi, and M.~Okumura, ``Generative replay inspired by hippocampal memory indexing for continual language learning,'' in \emph{Proceedings of the 17th Conference of the European Chapter of the Association for Computational Linguistics}, 2023, pp. 930--942.

\bibitem{ho2023prototype}
S.~Ho, M.~Liu, L.~Du, L.~Gao, and Y.~Xiang, ``Prototype-guided memory replay for continual learning,'' \emph{IEEE Transactions on Neural Networks and Learning Systems}, 2023.

\bibitem{schwarz2018progress}
J.~Schwarz, W.~Czarnecki, J.~Luketina, A.~Grabska-Barwinska, Y.~W. Teh, R.~Pascanu, and R.~Hadsell, ``Progress \& compress: A scalable framework for continual learning,'' in \emph{International conference on machine learning}.\hskip 1em plus 0.5em minus 0.4em\relax PMLR, 2018, pp. 4528--4537.

\bibitem{razdaibiedina2023progressive}
A.~Razdaibiedina, Y.~Mao, R.~Hou, M.~Khabsa, M.~Lewis, and A.~Almahairi, ``Progressive prompts: Continual learning for language models,'' \emph{arXiv preprint arXiv:2301.12314}, 2023.

\end{thebibliography}

\end{document}